\documentclass[12pt,a4paper,twoside]{article}
\usepackage[latin1]{inputenc}
\usepackage[T1]{fontenc}
\usepackage[francais]{babel}
\usepackage{times}
\usepackage{epsf}
\usepackage{amssymb}
\usepackage{Taln2003}

\begin{document}

\title{Une grammaire formelle du créole martiniquais pour la génération automatique}

\author{Pascal Vaillant\\
Université des Antilles-Guyane~--- UFR de Lettres et Sciences Humaines\\
B.P. 7207~--- 97275 {\sc Schoelcher cedex}~--- Martinique, France\\
Mél~: \texttt{pascal.vaillant@martinique.univ-ag.fr}}

\date{}

\maketitle

\motsClefs{Créole, Martiniquais, Grammaire, TAG, Génération}{Creole, Martiniquese, Grammar, TAG, Generation}

\section*{R\'esum\'e - Abstract}

Nous présenterons dans cette communication les premiers travaux de
modélisation informatique d'une grammaire de la langue créole
martiniquaise, en nous inspirant des descriptions fonctionnelles de
Damoiseau~\shortcite{Damoiseau1984} ainsi que du manuel de Pinalie \&
Bernabé~\shortcite{Pinalie1999}.  Prenant appui sur des travaux
antérieurs en génération de texte \cite{Vaillant1997}, nous utilisons
un formalisme de grammaires d'unification, les grammaires d'adjonction
d'arbres (TAG d'après l'acronyme anglais), ainsi qu'une modélisation
de catégories lexicales fonctionnelles à base syntaxico-sémantique,
pour mettre en \oe{}uvre une grammaire du créole martiniquais
utilisable dans une maquette de système de génération automatique.
L'un des intérêts principaux de ce système pourrait être son
utilisation comme logiciel outil pour l'aide à l'apprentissage du
créole en tant que langue seconde.

In this article, some first elements of a computational modelling of
the grammar of the Martiniquese French Creole dialect are presented.
The sources of inspiration for the modelling is the functional
description given by Damoiseau~\shortcite{Damoiseau1984}, and
Pinalie's \& Bernabé's~\shortcite{Pinalie1999} grammar manual.  Based
on earlier works in text generation \cite{Vaillant1997}, a unification
grammar formalism, namely Tree Adjoining Grammars (TAG), and a
modelling of lexical functional categories based on syntactic and
semantic properties, are used to implement a grammar of Martiniquese
Creole which is used in a prototype of text generation system.  One of
the main applications of the system could be its use as a tool
software supporting the task of learning Creole as a second language.

\pagebreak

\section{Introduction}

Le créole martiniquais est l'un des représentants de la famille des
langues créoles à base lexicale française de la zone Amérique-Caraïbes
(«~{\sc crelfacs}~»). Ces langues peuvent être décrites très
généralement parmi celles qui se sont formées dans une situation de
contact~; mais l'expression relève ici d'un euphémisme considérable,
le «~contact~» consistant en l'occurrence en la déportation massive de
peuples de langues différentes, obligés d'adopter en peu de temps un
véhiculaire comme nouveau moyen d'expression.  La situation historique
évoquée est celle de la colonisation du Nouveau Monde par les
puissances européennes entre le XVIIe et le XIXe siècle, et de
l'importation massive et forcée de main d'\oe{}uvre africaine
engendrée par le système économique esclavagiste.

Les créoles ont ainsi subi simultanément, en l'espace d'à peine
quelques générations, plusieurs processus d'évolution diachronique
rapides \cite{Alleyne1996}.  Certains sont assez faciles à se
représenter et sont l'objet d'une description relativement classique~:
ce sont par exemple les développements de tournures analytiques, du
même type que ceux qui ont conduit du latin au langues romanes
modernes, et qui sont encore à l'\oe{}uvre dans le français actuel
(plus développées dans les variétés de français populaire qui étaient
parlées par les colons que dans la langue standard, ces tournures
auraient ainsi pu selon Chaudenson~\shortcite{Chaudenson1995}
contribuer directement à influencer le lexique et la grammaire
créole).  D'autres, beaucoup plus spécifiques à la situation et
beaucoup moins à la portée de l'investigation linguistique historique,
ont pu provoquer des amalgames de structures lexicales et syntaxiques
françaises d'une part, et de structures provenant de plusieurs langues
africaines d'autre part, la propagation de modèles linguistiques
variant dans certains cas avec l'influence de certains groupes de
peuplement, qui jouissaient à certaines époques de situations de
domination démographique ou culturelle relative \cite{Alleyne1996}.
Les créoles ont en fin de compte développé un système sémantique et un
système syntaxique tout à fait original, tout en gardant
incontestablement une base lexicale majoritairement française.

Ces considérations historiques ne sont pas l'objet du travail présenté
ici, mais leur rappel permet de resituer dans leur perspective les
caractères originaux des crelfacs.

Dans une perspective synchronique, quoi qu'il en soit, nous nous
sommes attachés à mettre sur pied une modélisation adéquate et
économique des descriptions linguistiques du système actuel de la
langue. Nous nous sommes appuyés, pour le c\oe{}ur de la description
des énoncés, sur l'approche fonctionnelle de
Damoiseau~\shortcite{Damoiseau1984}, tout en ayant recours pour des
descriptions plus fines à certains exemples de \cite{Bernabe1983} et
de \cite{Pinalie1999}.

Le créole martiniquais est l'un des représentants de la famille des
crelfacs, avec le guyanais, le trinidadien (en voie d'extinction, au
profit de l'anglais), le saint-lucien, les dialectes dominicais, les
dialectes guadeloupéens, les dialectes haïtiens, et le louisianais. Il
a la particularité d'être l'un des plus unifiés, et partage par
ailleurs avec le créole de Port-au-Prince (devenu langue nationale
d'Haïti) une richesse en termes de production culturelle qui ne se
retrouve qu'à un degré moindre dans d'autres pays créolophones.
L'approche présentée ici est cependant aisément transposable à
d'autres langues du même groupe, notamment le guadeloupéen et le
guyanais.

\section{Description}

Nous exposons ici quelques faits grammaticaux succints~; des
descriptions plus complètes des faits de grammaire du créole
martiniquais se trouvent dans les ouvrages déjà cités
\cite{Damoiseau1984,Bernabe1983,Pinalie1999}.

Le créole est une langue isolante, dans laquelle les éléments lexicaux
sont invariables.  Les fonctions syntaxiques sont toutes manifestées
par l'ordre des éléments dans la phrase.

{\bf Groupes verbaux}

L'absence de flexion nominale est déjà familière à un lecteur
francophone.  Comme le français, et contrairement aux langues à
flexion nominale comme le latin ou le russe, le créole ne détermine la
fonction des actants et des circonstants des verbes que par deux
procédés~: (a) leur position dans la chaîne syntaxique, et (b) des
prépositions introduisant des cas sémantiques spécifiques.  L'ordre
des actants et des circonstants a tendance a être plus rigide encore
qu'en français, le procédé (a) ayant un périmètre d'emploi plus étendu
encore.  Ainsi, en créole, les verbes trivalents de type
«~attributif~» (correspondant à des verbes français comme {\em
donner}, {\em prêter}, {\em montrer}, etc.) forment des phrases à
l'aide d'un cadre de distribution syntaxique des actants qui comporte
trois groupes nominaux directs, là où en français seuls le sujet et
l'objet sont des groupes directs (le complément d'attribution étant
introduit par la préposition {\em à}\,).  Ceci impose un ordre strict,
en l'occurrence S V A O, pour déterminer les actants.  Ainsi, en
français les phrases (1a) et (1b) sont toutes deux recevables, alors
qu'en créole seule la phrase (2b) est recevable.

\begin{itemize}

\item[(1a)]{\em Pierre a donné un beau livre à Robert}

\item[(1b)]{\em Pierre a donné à Robert un beau livre}

\end{itemize}

\begin{itemize}

\item[(2a)]*{\em Pyè ba an bel liv Wobè}

\item[(2b)]{\em Pyè ba Wobè an bel liv}

\end{itemize}

La flexion verbale est ici remplacée par un système d'expression des
variations d'aspect et de temps qui est purement analytique, comme
celui que l'on observe en chinois.  Les différents degrés sont
exprimés par des particules temporelles ou aspectuelles qui sont
préposées au verbe (contrairement au chinois, où elles sont
postposées).

Le nombre de marques possibles du temps et de l'aspect est limité~; la
gamme d'emplois de chacune de ces marques est donc vaste.  Cet état de
choses fait de leur description sémasiologique une tâche considérable,
et il est difficile de caractériser le système de manière simple.  Un
panorama complet de cette question, ainsi qu'une description plus
particulière des différences entre créole guyanais et martiniquais,
est présenté par \cite{Pfaender2000}.

Plusieurs descriptions systémiques ont été proposées~; nous
retiendrons pour l'état actuel de notre modélisation celle de
Damoiseau \shortcite[p. 26]{Damoiseau1984}.  Si l'on peut questionner
la pertinence de la frontière stricte qu'il trace entre marques
temporelles et marques aspectuelles, notamment au sujet de la
classification de {\em ké} comme marque strictement aspectuelle, son
système présente l'avantage de recouper en assez grande partie la
combinatoire syntagmatique de ces marques~:

\begin{center}

\begin{tabular}{|l|c|c|c|}
\hline
{\sc temps~:} & \multicolumn{3}{c|}{\sc aspect~:} \\
\hline
{~} & perfectif & imperfectif & imperfectif \\
{~} & {~} & (duratif, général, itératif) & (prospectif) \\
\hline
(pas d'indication) & $\varnothing{}$ & ka & ké \\
\hline
passé & té & té ka & té ké \\
\hline
\end{tabular}

\end{center}

Les combinaisons $\varnothing{}$, {\em ka}, {\em ké}, {\em té}, {\em
té ka} et {\em té ké} sont en effet les plus fréquemment
attestées\footnote{Les choses sont plus complexes en réalité, puisque
la combinaison {\em ké ka} est également attestée avec une valeur de
futur duratif~--- cf. \cite[p. 77]{Pinalie1999}~: {\em lè ou ké rivé,
nou ké ka dòmi}~: quand vous arriverez, nous serons en train de
dormir.  La combinaison {\em té ké ka} est même possible (comme dans
la chanson de zouk~: «~{\em Kolé séré nou té ké ka dansé}~»~$\simeq{}$
nous serions en train de danser l'un contre l'autre).
Valdman~\shortcite[p. 219]{Valdman1978} expose un système plus complet
fondé sur deux catégories d'aspect ($\pm{}$continuatif,
$\pm{}$prospectif), et une catégorie de temps ($\pm{}$passé).}.

Pour donner une idée des valeurs de ces combinaisons de marques, on
peut mentionner certaines de leurs traductions les plus fréquentes en
français~: {\em mwen dòmi~:} j'ai dormi~; {\em mwen ka dòmi~:} je
dors~; {\em mwen ké dòmi~:} je vais dormir~; {\em mwen té dòmi~:}
j'avais dormi~; {\em mwen té ka dòmi~:} je dormais~; {\em mwen té ké
dòmi~:} j'aurais dormi.  Il faut cependant bien entendu garder à
l'esprit qu'il ne s'agit pas d'équivalences systématiques, puisque les
systèmes aspecto-temporels ne se recoupent pas, et que la notion de
«~mode~» est étrangère au créole.

En première approche, on peut schématiser les domaines d'application
des systèmes aspectuel et temporel en disant que la caractérisation de
l'aspect ($\varnothing{}$, {\em ka} ou {\em ké}\,) ne porte que sur des
verbes exprimant un procès, alors que la caractérisation du temps
($\varnothing{}$ ou {\em té}\,) s'applique aussi bien aux procès qu'aux
états.  On a donc deux séries de verbes, celle qui peut suivre les
marques $\varnothing{}$, {\em ka}, {\em ké}, {\em té}, {\em té ka} et
{\em té ké}, et celle qui ne peut suivre que $\varnothing{}$ ou {\em
té}.  C'est en ces deux séries de verbes, par exemple, que
\cite{SaintQuentin1872}, l'une des premières grammaires créoles,
voulait distinguer deux «~conjugaisons~» du créole guyanais.

Selon cette description, les phrases (3a) et (3b) sont donc possibles
et leur différence traduit une nuance d'aspect, alors que seule la
phrase (4a) est recevable.

\begin{itemize}

\item[(3a)]{\em I dòmi tout lajounen} (il a dormi toute la journée)

\item[(3b)]{\em I ka dòmi tout lajounen} (il passe ses journées à dormir)

\end{itemize}

\begin{itemize}

\item[(4a)]{\em I ni anpil lajan} (il a beaucoup d'argent)

\item[(4b)]*{\em I ka ni anpil lajan}

\end{itemize}

Cette description est en réalité une simplification, puisqu'il existe
bel et bien des contextes dans lesquels l'application de marques
aspectuelles à des verbes «~à signifié non-processif~»
\cite{Damoiseau1984} est attestée~: {\em ravèt pa ka ni rézon douvan
poul} (le ravet n'a jamais raison devant la poule)~; ou même~: {\em lè
i ka sòti kazino i ka ni anpil lajan}~: à chaque fois qu'il sort du
casino, il a plein d'argent.  Nous nous en tiendrons cependant à la
première approche dans le cadre de ces premières tentatives de
modélisation.

{\bf Groupes nominaux}

Les groupes nominaux du créole ont un système de degrés de
détermination qui comprend un degré générique, exprimé par
$\varnothing{}$ (ainsi {\em ravèt} et {\em poul} dans l'exemple
ci-dessus), un degré indéterminé ({\em an timanmay}~: un enfant), un
degré défini exprimé par un déterminant postposé ({\em timanmay-la}~:
l'enfant), et un degré démonstratif également exprimé par un
déterminant postposé ({\em timanmay-tala}~: cet enfant).  Les
déterminants postposés, en martiniquais, peuvent se combiner avec une
marque préposée {\em sé-} pour former un pluriel défini ({\em
sé-timanmay-tala}~: ces enfants)~--- le pluriel générique du français
n'ayant pas d'équivalent~: on utilise le degré générique du singulier.

Il n'existe pas de genre grammatical~; en revanche les noms
déterminent une variation dans le choix de l'article défini, selon un
principe d'harmonie consonantique et nasale~: les mots se terminant
par une voyelle ouverte prennent la marque {\em -a} comme déterminant
défini, les mots se terminant par une consonne ou par une semi-voyelle
({\em y} ou {\em w}\,) prennent la marque {\em -la}.  Ceci ne s'applique
qu'aux mots dont la dernière syllabe est non-nasalisée~; Pour les mots
à terminaison nasale, les marques du défini sont respectivement {\em
-an} et {\em -lan}.

{\bf Parties du discours}

L'un des faits les plus frappants du créole, enfin, est l'absence de
frontières nettes de catégories grammaticale.  La tentation de
rapprocher les mots d'origine française de leurs étymons et de leur
attribuer la même catégorie (nom, verbe, adjectif ...) est souvent
grande, et elle a même assez fréquemment une certaine pertinence, mais
cette analogie est trompeuse.  Dans beaucoup de cas en effet, si un
mot peut être utilisé dans le même emploi grammatical que son étymon
français, il possède également d'autres emplois, propres à des
catégories syntaxiques différentes.

Les adjectifs, par exemple, ont en français besoin d'un verbe copule
pour figurer en fonction d'attribut~: {\em le gros arbre} vs. {\em
l'arbre est gros}.  En créole, l'adjectif seul peut jouer le rôle de
prédicat~: {\em gwo pyébwa-a} vs. {\em pyébwa-a gwo} (cf. chinois {\em
dà shùmù} vs. {\em shùmù dà}\,).  L'adjectif dans cette position de
prédicat-attribut peut également recevoir les marques temporelles de
la même manière que les verbes d'état~: {\em pyébwa-tala té gwo}~: cet
arbre était gros~; voire des marques aspectuelles dans certains
contextes, comme les verbes d'état également (exemple cité par
Damoiseau~\shortcite[p. 29]{Damoiseau1984}~: {\em i ka las vit}~: il
se fatigue vite).  Il est en fait très difficile de fournir une
définition permettant de distinguer le verbe en tant que catégorie
distincte de l'adjectif autrement que par des critères négatifs
(blocage de la fonction épithète).

Au-delà du cas des verbo-adjectivaux, des «~noms~» peuvent apparaître
en position de prédicat ({\em i yich mwen}~: c'est mon enfant), et des
«~verbes~» en fonction nominale ({\em mwen dòmi an bel dòmi}~: j'ai
sacrément dormi $[$litt. j'ai dormi un grand dormir$]$~; {\em vin pran
an bwè}~: viens prendre un verre $[$litt. viens prendre un boire$]$)~;
et ces tournures sont loin d'être des exceptions ou des figures de
style, mais font partie du fonctionnement le plus banal de la langue.
Certains verbes, en outre, occupent des fonctions de préposition.
Ainsi, le verbe {\em ba} (donner), qui construit généralement des
phrases avec trois actants ({\em i {\bf ba} mwen an pen}~: il m'a
donné un pain), peut, accompagné du seul actant {\em destinataire},
construire un syntagme qui joue le rôle de circonstant pour les verbes
n'ayant pas le destinataire dans leur cadre valenciel de base. Ainsi
dans {\em i pòté an boutèy wonm {\bf ba} mwen}~: il m'a apporté une
bouteille de rhum, ou {\em mwen ka palé {\bf ba}'w}~: je suis en train
de te parler (cf. chinois {\em t\={a} {\bf g\v{e}i} w\v{o} yi píng
ji\v{u}}~: il me donne une bouteille d'alcool, vs. {\em t\={a} dàilái
le yi píng ji\v{u} {\bf gei} w\v{o}}, il m'a apporté une bouteille
d'alcool).  Certains verbes enfin peuvent être employés comme
particules aspectuelles, enrichissant le système fondé sur
\{$\varnothing$, {\em ka}, {\em ké} et {\em té}\}~; ainsi {\em sòti}
(sortir) ou {\em fini} (finir) dans des tournures comme~: {\em es ou
kompwann sa yo sòti di'w la~?}~: est-ce que tu comprends ce qu'ils
viennent de te dire~? \cite[p. 1057]{Bernabe1983}, ou {\em pwan}
(prendre) dans~: {\em i pwan kouri}~: il s'est mis à courir
\cite[p. 1043]{Bernabe1983}.

\section{Modélisation}

On pourrait donner encore d'innombrables exemples de cette plasticité
grammaticale~; la leçon qu'il convient d'en tirer dans le cadre de ce
travail (modélisation informatique), est qu'il faut développer un
mécanisme pour la prendre en compte, car elle est tout sauf marginale.

Il serait peu pratique de multiplier les catégories grammaticales
{\em ad hoc} en faisant un produit cartésien sur les catégories de la
grammaire française pour obtenir des parties du discours
sur-spécialisées (adjectif-verbe-qui-ne-peut-jamais-être-un-nom, ou
nom-pouvant-servir-parfois-de-verbe-d'état~...).  Il nous paraît
également qu'il serait assez artificiel de prévoir un grand nombre de
règles de conversion~; il semble en définitive que la catégorie
grammaticale soit une notion moins essentielle en créole martiniquais
qu'en français, ou autrement dit~--- car il faut bien tout de même des
fonctions dans l'énoncé~--- qu'elle soit plus qu'ailleurs
contextualisable.  Le fait qu'un mot ait la possibilité de remplir une
fonction de sujet (comme un groupe nominal), d'épithète (comme un
adjectif), de prédicat (comme un verbe)~... semble pouvoir être
déterminé de manière afférente aussi bien qu'inhérente, pour reprendre
ici le vocabulaire de Rastier~\shortcite{Rastier1987}.  Un mot peut
avoir un emploi central de verbe, mais, placé dans un contexte de
groupe nominal (avec un article, par exemple), pouvoir être interprété
sans difficulté comme un nom.  De même, un nom, placé directement
après le sujet, et éventuellement temporalisé par une particule comme
{\em té}, peut jouer le rôle d'un verbe si son sens s'y prête.  Ces
réflexions sont d'ailleurs probablement valides {\em en principe} pour
toutes les langues\footnote{Ainsi, un énoncé comme {\em je mère-poule
mes enfants} est tout à fait compréhensible~--- sinon recevable~--- en
français (exemple emprunté à Jacques Coursil).}, quoiqu'à des degrés
variables.

Il a donc été choisi de ne retenir que deux grandes catégories
grammaticales ({\bf N} et {\bf Pred}~: nom et prédicat), lesquelles
peuvent être déterminées plus spécifiquement par des ensembles de
traits caractérisant leurs emplois possibles~:

{\bf Pred} possède les attributs $<$\texttt{type}$>$\texttt{=T},
$<$\texttt{cadre}$>$\texttt{=C}, $<$$\pm{}$\texttt{epithete}$>$. Le
{\em type} T classe le mot soit parmi les verbes exprimant un procès
({\em dòmi}~: dormir, {\em manjé}~: manger, {\em ba}~: donner, {\em
pòté}~: porter, etc.), soit parmi les verbes ou les verbo-adjectivaux
exprimant un état ({\em ni}~: avoir, {\em sav}~: savoir, {\em gwo}~:
gros/être gros, {\em las}~: fatigué/être fatigué, etc.).  Le {\em
cadre} C correspond au cadre d'attribution de positions syntaxiques
aux actants du verbe (sous-catégorisation verbale)~: intransitif pour
{\em dòmi} ou {\em las}, transitif pour {\em manjé}, attributif pour
{\em ba}, etc.  Le trait {\em épithète}, à valeur binaire, détermine
si le mot peut oui ou non apparaître en position épithète, comme un
adjectif~: il est positif pour {\em gwo} et négatif pour {\em manjé},
par exemple.

{\bf N} possède l'attribut $<$\texttt{harm}$>$=\texttt{A}, qui
détermine le type de marque de détermination définie qui doit
apparaître selon la syllabe finale (A peut valoir {\em a}, {\em la},
{\em an} ou {\em lan}\,).  Les noms propres et les pronoms personnels
possèdent également le trait $<$\texttt{det}$>$\texttt{=def}, qui
exprime le fait qu'ils sont déjà définis (et ne se combinent donc plus
avec un déterminant)~--- le trait $<$\texttt{det}$>$ n'apparaissant
dans les autres cas qu'au niveau de barre 2, c'est-à-dire au niveau du
groupe nominal.  Le fait qu'un {\bf N} puisse oui ou non apparaître
également en fonction verbale est déterminé par la présence, à côté de
ces traits spécifiquement nominaux, du trait
$<$\texttt{cadre}$>$\texttt{=C} signalant l'existence d'une valence
interne.

Cette conception des catégories comme faisceaux de traits évoque celle
qui guide le formalisme des grammaires syntagmatiques généralisées
(GPSG) \cite[p. 101--105]{Abeille1993}.  Nous avons toutefois préféré
utiliser pour ce travail le formalisme des grammaires d'arbres
adjoints à unification de structures de traits (FS-TAG)
\cite{VijayShankerJoshi1988}, \cite[ch. 4]{Abeille1993}, qui se prête
tout à fait bien à la génération (bien qu'il ait d'abord été conçu
pour l'analyse) grâce à sa capacité à représenter les tours de phrase
sous la forme d'arbres élémentaires s'étendant sur plus d'un seul
niveau.  L'utilisation de modèles d'arbres adjoints pour la génération
a été pour cette raison adoptée dans différents travaux de recherche,
d'autant qu'elle a pu être étendue au-delà du niveau de la phrase,
comme dans le modèle G-TAG \cite{Danlos1998}.

Nous n'avons en revanche pas, pour cette application de génération,
une grammaire TAG entièrement lexicalisée~; les arbres sans ancre
lexicale nous permettent en effet de donner une description formelle
homogène à des structures parallèles, dont les unes s'expriment avec
une marque lexicale et les autres avec une marque zéro
(ex. imperfectif en {\em ka} vs. perfectif en $\varnothing{}$).  En
outre, les arbres sans ancre lexicale permettent (a)~de faire
l'économie de quelques kilo-octets de fichier, et de nombreuses
opérations de copie-colle, en regroupant sous un même «~arbre schéma~»
toutes les structures verbales ayant le même cadre de
sous-catégorisation~; (b)~de rendre compte des phénomènes
linguistiques de degré de complexité supérieur à la phrase nucléaire
(comme les subordonnées relatives) \cite[p. 230]{Abeille1993}.

{\bf Construction du GN}

\begin{figure}[htb]
 \begin{center}
 \makebox[159mm][c]{\parbox[t]{50mm}{\epsffile{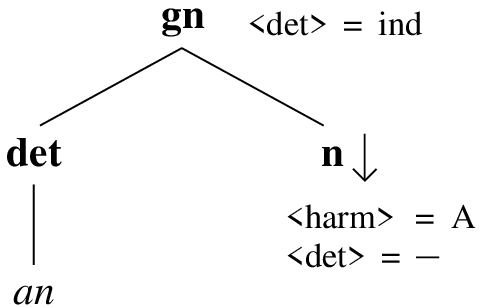}}
                    \parbox[t]{45mm}{\epsffile{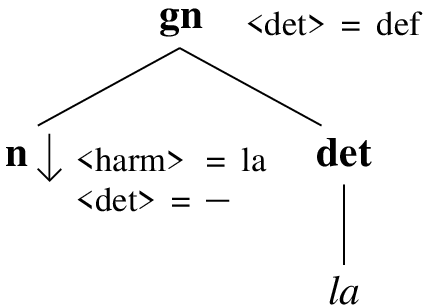}}
                    \parbox[t]{30mm}{\epsffile{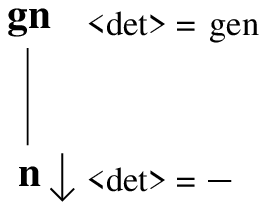}}
                    \parbox[t]{30mm}{\epsffile{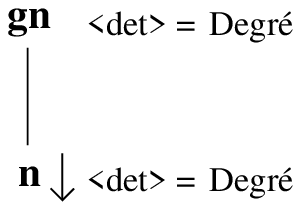}}}
 \leavevmode
 \caption[GN-det]{Arbres élémentaires~: (a) du déterminant indéfini~;
                  (b) d'un déterminant défini~; (c) du degré de
                  détermination générique~; (d) du nom déjà défini.}
 \label{GN-det}
 \end{center}
\end{figure}

Le {\bf N} non encore défini, mais qui peut avoir déjà reçu des
compléments (en termes de théorie de X-barre, le {\bf \={N}}), se
substitue à la position adéquate dans l'arbre élémentaire du
déterminant pour former le {\bf GN} (niveau de barre 2).  Le cas du
déterminant indéfini est illustré fig.~\ref{GN-det}\/a (la valeur du
trait {\em harm}, A, est une variable), et un exemple de déterminant
défini est illustré fig.~\ref{GN-det}\/b\footnote{N.B. Dans ces
figures, comme dans les suivantes, conformément à la tradition héritée
du langage Prolog et du formalisme des grammaires à clauses définies
(DCG), les termes commençant par une majuscule correspondent à des
variables, alors que les termes commençant par des minuscules
correspondent à des valeurs instanciées.}.

Cette modélisation place le déterminant défini postposé {\em après}
les éventuels compléments, comme les compléments de nom, ou les
subordonnées relatives.  Ceci est conforme avec l'usage du créole
martiniquais, qui dit~: {\em kay-la} (la maison)~; {\em kay papa mwen
an} (la maison de mon père)~; {\em kay mwen ja palé ba'w la} (la
maison dont je t'ai parlé).

Le {\bf GN} déterminé au degré générique se construit à partir du nom
non encore déterminé grâce à l'arbre représenté en
fig.~\ref{GN-det}\/c~; tandis que le {\bf GN} peut également se
constituer directement sur la base d'un {\bf N} pré-déterminé au
niveau lexical, comme un pronom personnel ou un nom propre, comme
illustré fig.~\ref{GN-det}\/d.

{\bf Construction du GPred}

Le prédicat connaît trois niveaux d'expansion, conformément au choix
que nous avons fait de nous conformer au modèle simplifié du temps et
de l'aspect présenté par Damoiseau~\shortcite{Damoiseau1984} (cf. plus
haut).

\begin{figure}[htb]
 \begin{center}
 \makebox[159mm][c]{\parbox[t]{45mm}{\epsfxsize=40mm\epsffile{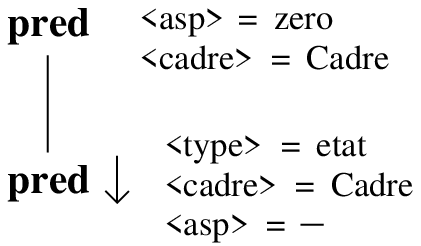}}
                    \parbox[t]{45mm}{\epsfxsize=40mm\epsffile{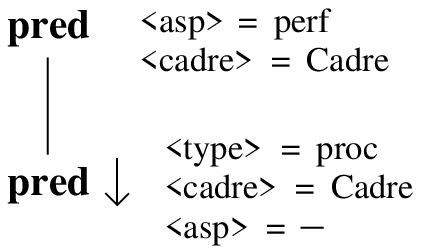}}
                    \parbox[t]{65mm}{\epsfxsize=60mm\epsffile{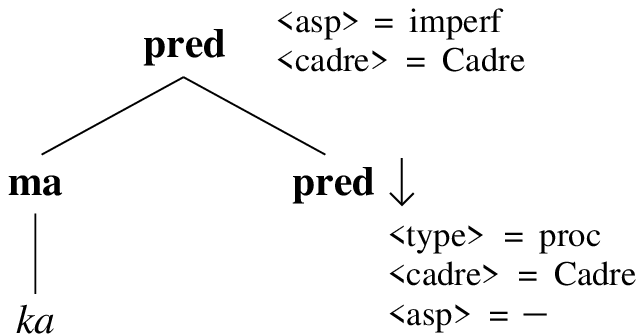}}}
 \leavevmode
 \caption[Pred-Asp]{Arbres de construction du prédicat~: (a) aspect zéro~; (b) aspect perfectif~; (c) aspect imperfectif.}
 \label{Pred-Asp}
 \end{center}
\end{figure}

Au niveau lexical, le {\bf Pred} n'est marqué ni en temps ni en
aspect.  Au niveau 1, il est marqué en aspect~: soit avec une valeur
d'aspect {\em zéro}, dans le cas où il s'agit d'un état~; soit avec
une valeur d'aspect {\em perfectif}, {\em imperfectif} ou {\em
prospectif}, s'il s'agit d'un procès.  Les arbres correspondant à
l'aspect zéro (pour un état), et à l'aspect perfectif (pour un
procès), sont montrés respectivement en fig.~\ref{Pred-Asp}\/a et
fig.~\ref{Pred-Asp}\/b.  Les aspects imperfectif et prospectif sont
formés à l'aide des marques {\em ka} et {\em ké}.  L'arbre
correspondant au cas de l'aspect imperfectif est montré en
fig.~\ref{Pred-Asp}\/c.  L'aspect prospectif se construit de la même
manière.

Au niveau 2, le prédicat est marqué en temps, soit avec la marque zéro
($\varnothing{}$), soit avec la marque du passé ({\em té}\,)~; les
mécanismes utilisés correspondent respectivement, à un niveau
d'expansion au-dessus, à ceux utilisés pour marquer respectivement
l'aspect perfectif ou l'aspect imperfectif sur les prédicats-procès.
La marque de temps s'applique à un terme de niveau 1 ($\bar{\mbox{\rm
Pred}}$) qui peut donc être soit un état (automatiquement à l'aspect
zéro), soit un procès (déjà marqué en aspect).  Dans l'arbre
élémentaire servant à passer du prédicat marqué en aspect
($\bar{\mbox{\rm Pred}}$) au prédicat marqué en temps et en aspect
($\bar{\bar{\mbox{\rm Pred}}}$), le n\oe{}ud inférieur (n\oe{}ud de
substitution) doit donc spécifier la présence du trait {\em aspect} et
l'absence du trait {\em temps}, puis reporter le trait {\em aspect} au
niveau supérieur, et y ajouter le trait {\em temps}~--- le n\oe{}ud
supérieur comportant donc en tout les traits {\em temps}, {\em aspect}
et {\em cadre}.

Enfin, au niveau supérieur, {\bf GPred}, le prédicat marqué en temps
et en aspect ($\bar{\bar{\mbox{\rm Pred}}}$) est entouré de ses
actants, dans la disposition spécifique du cadre de
sous-catégorisation concerné.  Le trait {\em cadre} est supprimé, et
un trait {\em saturé} est ajouté. La phrase nucléaire ({\bf Ph}) est
constituée d'un sujet et d'un {\bf GPred} saturé.

\begin{figure}[htb]
 \begin{center}
 \makebox[55mm][l]{\epsfxsize=55mm\epsffile{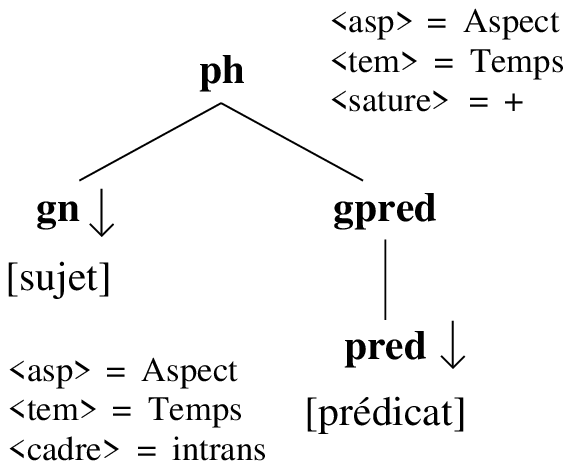}}
 \makebox[95mm][r]{\epsfxsize=65mm\epsffile{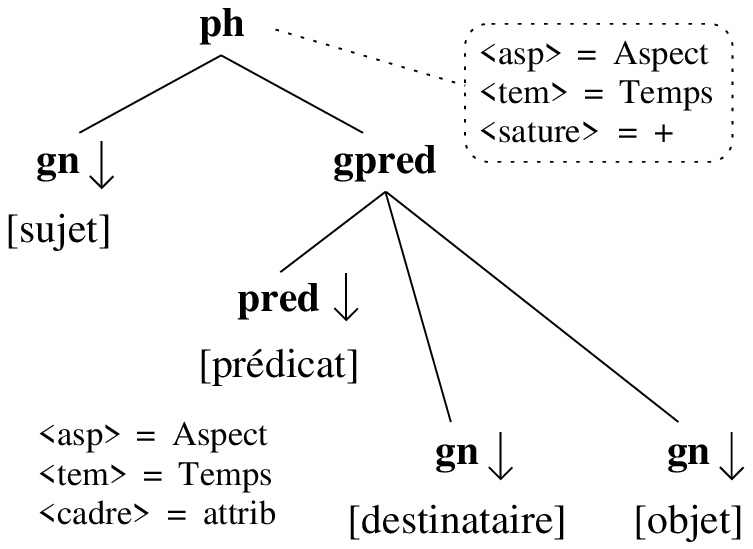}}
 \leavevmode
 \caption[Arbres-Schema]{Arbres schéma~: (a) des verbes intransitifs~; (b) des verbes attributifs.}
 \label{Arbres-Schema}
 \end{center}
\end{figure}

Les différents cadres de sous-catégorisation sont représentés par des
arbres-schéma (fig.~\ref{Arbres-Schema}), qui engendrent des familles
d'arbre lorsque l'on y greffe les arbres des différents prédicats
marqués en temps et en aspect.

\addtolength{\belowcaptionskip}{-2mm}

\begin{figure}[htb]
 \begin{center}
 \makebox[100mm][c]{\epsfxsize=65mm\epsffile{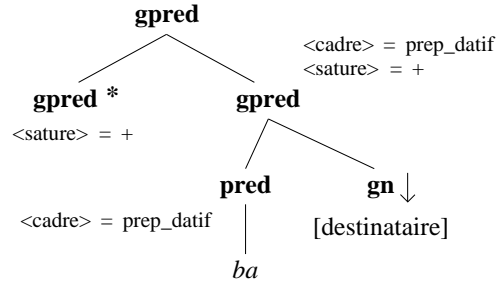}}
 \leavevmode
 \caption[Arbre-Aux-Schema]{Arbre auxiliaire pour exprimer les compléments d'attribution.}
 \label{Arbre-Aux-Schema}
 \end{center}
\end{figure}

Les circonstants, c'est-à-dire les rôles sémantiques qui n'ont pas pu
être exprimés par les actants du cadre valenciel de base du verbe,
sont exprimés par l'adjonction de syntagmes prépositionnels.  Comme
dans le cas déjà évoqué des compléments d'attribution introduits par
{\em ba}, le rôle des prépositions peut parfois être joué par des
verbes accompagnés d'un jeu «~restreint~» d'actants.  Ces types de
syntagmes sont modélisés comme des {\bf GPred}, à la différence que le
verbe y est utilisé avec un autre cadre de sous-catégorisation~; le
cas est donc traité comme une alternance verbale.  Ainsi, le verbe
{\em ba} possède, à côté de son cadre «~complet~» ({\em attributif}),
un cadre «~restreint~» ({\em prep\_datif}) qui est utilisé pour former
les arbres auxiliaires pour l'adjonction de compléments d'attribution,
grâce à l'arbre-schéma représenté fig.~\ref{Arbre-Aux-Schema}.  Un
trait {\em saturé}, contraignant l'adjonction sur le n\oe{}ud pied de
ce type d'arbres, garantit que les circonstants ne peuvent commencer à
s'adjoindre au {\bf GPrep} qu'après tous les actants (sujet excepté).

\section{Mécanisme de génération}

Le programme réalisé implante informatiquement les arbres décrits par
des structures {\sc prolog}.  Toutes les opérations élémentaires sur
les structures de traits (unification, écrasement) ont été également
implantées sous la forme de buts {\sc prolog}, ainsi que les
opérations sur les arbres spécifiques au modèle TAG (substitution et
adjonction).

La structure de départ du processus de génération est un graphe
conceptuel contenant des n\oe{}uds liés entre eux par des rôles
sémantiques.  Les n\oe{}uds peuvent porter des attributs sémantiques
qui servent à en préciser la valeur (temps, aspect, degré de
détermination~...)  Le programme effectue la génération en parcourant
le graphe conceptuel en un parcours direct (arbre de couverture du
graphe), et en engendrant, pour chaque prédicat, un arbre
correspondant à la verbalisation de ce prédicat et de ses actants
fondamentaux sous la forme d'une phrase nucléaire.  Les rôles
sémantiques non-verbalisés dans la phrase nucléaire sont ensuite
générés sous forme d'arbres auxiliaires à tête {\bf GPred}, et
adjoints à la phrase nucléaire.  Enfin, les relations sémantiques
incidentes à l'un des concepts de l'arbre de couverture direct, mais
non encore exprimées dans la phrase principale, sont soit adjointes
lorsque c'est possible par un procédé syntaxique commun (épithète,
subordonnée relative), soit générées dans de nouvelles phrases
juxtaposées à celles déjà engendrées.  Ce mécanisme a été exposé plus
en détail dans \cite{Vaillant1997,Vaillant1998}, et reste
fondamentalement le même, indépendamment de la langue à laquelle il
est appliqué.

\section{Conclusion}

Dans ce travail, une grammaire à adjonction d'arbres a été utilisée
pour passer d'une structure sémantique limitée au niveau phrastique à
une lexicalisation en langue créole. Nous n'avons pas eu l'ambition en
revanche d'essayer de partir d'une structure sémantique
«~interlangue~» ou «~indépendante de la langue~».  Le système des
paramètres sémantiques utilisés pour le temps et l'aspect, par
exemple, est ni plus ni moins celui de la langue créole martiniquaise
(ou tout ou moins du sous-ensemble que nous prétendons en modéliser).
La question de l'existence d'une étape encore antérieure à la
représentation liée à la structure sémantique de la langue-cible n'est
pas abordée dans le cadre de cette étude, qui se borne à réaliser la
génération de la chaîne syntagmatique à partir d'un modèle du sens
dans la même langue, représenté sous forme de graphe.  Tout au plus
peut-on noter que le même système peut servir, à quelques
modifications de la grammaire près, pour des dialectes proches, qui
ont à peu près le même système sémantique (ainsi le guadeloupéen, ou
le guyanais de la côte).

La question de l'étape antérieure se pose en réalité dans deux
contextes~: soit dans les tentatives de modélisation des processus
cognitifs fondamentaux, qui postulent l'existence d'une hypothétique
représentation sémantique profonde, ou universelle~; soit dans les
applications de transfert d'un système sémiotique à un autre, donc de
génération de phrases créoles, par exemple, à partir d'une entrée qui
pourrait aussi bien être formulée en français, en allemand, en
chinois, ou en idéogrammes Bliss.  Nous viendrons à l'avenir à nous
pencher sur le second problème, qui pourra être le noyau d'un futur
système d'aide à l'apprentissage, destiné par exemple à des apprenants
n'ayant pas le créole pour langue maternelle.  Quant au premier
problème, il nous semble hors de notre portée.  On pourra se consoler
en évoquant l'hypothèse du créoliste innéiste
Bickerton~\shortcite{Bickerton1981}, selon lequel ressurgissent dans
les langues créoles les structures innées du langage faisant partie du
«~bioprogramme~» de l'humanité.  Selon cette hypothèse, donc, s'il
existe une quelconque structure sémantique universelle, c'est
précisément celle que nous avons sous la main avec le créole~...

\bibliographystyle{taln2003}
\bibliography{Creole}

\end{document}